\documentclass[10pt,twocolumn,letterpaper]{article}

\usepackage[algorithms]{wacv}      

\usepackage{graphicx}
\usepackage{amsmath}
\usepackage{amssymb}
\usepackage{booktabs}

\usepackage[pagebackref,breaklinks,colorlinks]{hyperref}

\usepackage{multirow}
\usepackage{makecell}
\usepackage{pifont}
\newcommand{\cmark}{\ding{51}}
\newcommand{\xmark}{\ding{55}}

\usepackage[capitalize]{cleveref}
\crefname{section}{Sec.}{Secs.}
\Crefname{section}{Section}{Sections}
\Crefname{table}{Table}{Tables}
\crefname{table}{Tab.}{Tabs.}

\def\wacvPaperID{60} 
\def\confName{WACV}
\def\confYear{2025}

\begin{document}

\title{The Linear Attention Resurrection in Vision Transformer}

\author{Chuanyang Zheng\\
{\tt\small chuanyang\_zheng@sjtu.edu.cn}\\
\url{https://github.com/ChuanyangZheng/L2ViT}
}
\maketitle

\begin{abstract}
	Vision Transformers (ViTs) have recently taken computer vision by storm. However, the softmax attention underlying ViTs comes with a quadratic complexity in time and memory, hindering the application of ViTs to high-resolution images. We revisit the attention design and propose a linear attention method to address the limitation, which doesn't sacrifice ViT's core advantage of capturing global representation like existing methods (\eg local window attention of Swin). We further investigate the key difference between linear attention and softmax attention. Our empirical results suggest that linear attention lacks a fundamental property of concentrating the distribution of the attention matrix. Inspired by this observation, we introduce a local concentration module to enhance linear attention. By incorporating enhanced linear global attention and local window attention, we propose a new ViT architecture, dubbed L$^2$ViT. Notably, L$^2$ViT can effectively capture both global interactions and local representations while enjoying linear computational complexity. Extensive experiments demonstrate the strong performance of L$^2$ViT. On image classification, L$^2$ViT achieves 84.4\% Top-1 accuracy on ImageNet-1K without any extra training data or label. By further pre-training on ImageNet-22k, it attains 87.0\% when fine-tuned with resolution 384$^2$. For downstream tasks, L$^2$ViT delivers favorable performance as a backbone on object detection as well as semantic segmentation. 
\end{abstract}

\section{Introduction}
The computer vision community has witnessed the prosperity of convolutional neural networks (CNNs) \cite{he2016deep, krizhevsky2012imagenet, tan2019efficientnet} over the last decade. Recently, vision transformers rise rapidly and have yielded impressive performances on various vision tasks including image classification~\cite{liu2021swin}, object detection~\cite{dong2022cswin}, segmentation~\cite{strudel2021segmenter} and so on. Beginning with the pioneering work of ViT \cite{dosovitskiy2020image}, which first challenges CNNs with the vanilla transformer on image classification, ViTs have evolved to become increasingly powerful. The key component behind the success of ViTs is self-attention, which empowers ViTs with a global receptive field, adaptive data specificity, and more human-like representations~\cite{park2021vision,tuli2021convolutional}.

These advantages, however, come with quadratic computational complexity in time and memory with respect to input resolution. Various methods are proposed to address this issue and make ViTs applicable in more downstream tasks such as object detection. The first representative approach is to restrict the softmax attention to fixed-size window ranges, such as local 7x7 window~\cite{liu2021swin}, sliding window~\cite{zhu2021long}. However, this line of work has been observed to have limited model capacity due to the sacrifice of the global receptive field, which brings strong model capacity~\cite{ding2022scaling}.
Another typical approach aims to reduce the number of keys or values in attention via linear projection~\cite{wang2021pyramid}, convolution~\cite{zhang2022rest}, pooling~\cite{fan2021multiscale}. When targeting high-resolution input in dense prediction tasks, they apply a relatively large downsampling ratio in the earlier stages, \eg, 8 in the first stage, to reduce the computational cost. This will inevitably damage the model's performance since aggressive downsampling operations lose some crucial context information and destroy the global dependency modeling ability of self-attention to a certain extent.

\begin{figure*}[t]
	\begin{center}
		\includegraphics[width=1\textwidth]{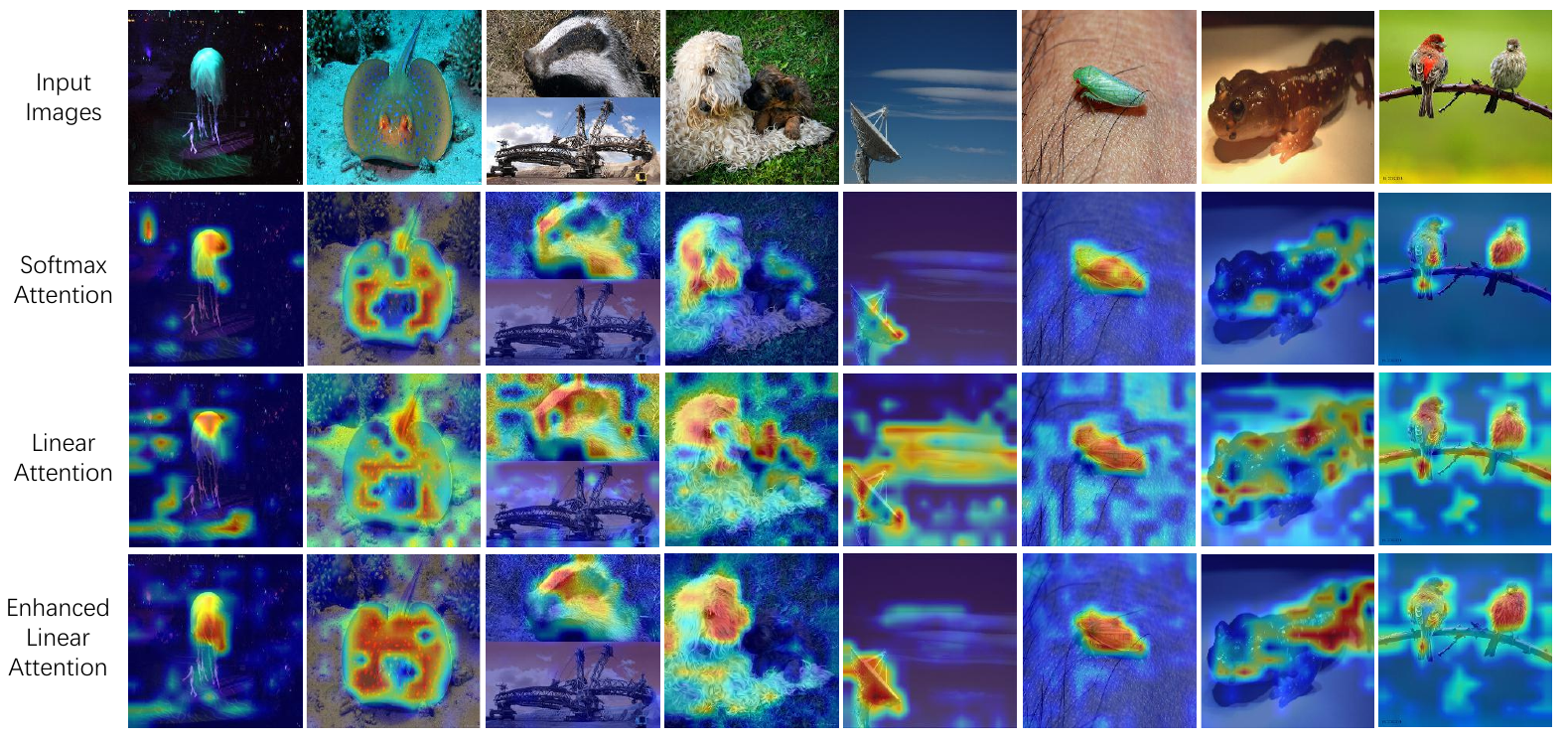}
	\end{center}
	\caption{Grad-CAM~\cite{selvaraju2017grad} activation maps of DeiT-Tiny~\cite{touvron2021training} equipped with different attention mechanisms, i.e., softmax attention, linear attention, and enhanced linear attention. The first row is the original input images. Enhanced linear attention can substantially eliminate some irrelevant distractions and focus better on the object itself, such as the objects in the fifth and last columns.}
	\label{fig:grad-cam}
\end{figure*}

To overcome the above issues, we propose to replace softmax attention with linear attention~\cite{katharopoulos2020transformers,cai2023efficientvit}. In this paper, linear attention refers to the kernel-based attention mechanism, detailed in the related work, and not all attention variants are with linear complexity.
On the one hand, linear attention takes advantage of the associativity property of matrix products to achieve computational complexity of $O(N)$ ($N$ is the number of patches in vision transformers). On the other hand, linear attention still models communications among all tokens and learns a global spatial relationship, which is essential for visual recognition tasks and hurt by the above attention variants. Nevertheless, previous works~\cite{hong2022fair,liu2021swin,rao2022amixer,qin2021cosformer,zhang2021multi} show linear attention performs inferiorly compared to other attention variants in vision transformer. We thoroughly investigate softmax attention and linear attention, demystifying two key insights of softmax attention. The first property is that all values in the attention map must be non-negative as verified in~\cref{tab:non-negative}, so we apply ReLU as a feature mapping function to guarantee non-negative attention values.~\cref{fig:non-negative} suggests ReLU-based linear attention can capture a similar relationship as vanilla attention. The second property is the concentration of attention in vanilla ViT (the second row in~\cref{fig:grad-cam}). Without re-weighting the attention matrix by softmax, linear attention fails to concentrate on crucial local information (the third row in~\cref{fig:grad-cam}). Thus we introduce a local concentration module to improve linear attention (the last row in~\cref{fig:grad-cam}).

Though enhanced linear attention learns global interactions effectively, local information is less preserved. To further strengthen locality, we propose a new general-purpose backbone named L$^2$ViT (\textbf{L}inear global attention and \textbf{L}ocal window attention \textbf{Vi}sion \textbf{T}ransf-ormer). L$^2$ViT integrates the enhanced linear attention and local window self-attention in an alternatively sequential way as shown in~\cref{fig:overall}. The local window self-attention introduces locality and translational invariance that have been proven beneficial for vision tasks, making L$^2$ViT better at modeling fine-grained and short-distance representations. Instead, linear attention maintains long-range dependency and constructs a global context-rich representation from the whole image, providing a large effective receptive field. The alternative design mixes these complementary feature information and provides powerful modeling capacity with only linear complexity. 

The proposed L$^2$ViT architecture demonstrates effectiveness on a broad spectrum of
vision tasks, ranging from image classification to objection detection and semantic segmentation. 
Furthermore, pre-training on more data and equipping with common model augmentation strategies can push L$^2$ViT to achieve stronger performance. With these encouraging results, we hope L$^2$ViT can provide useful insights for further research in visual recognition.

\section{Related Work}
\subsection{Vision Transformer}
Although the tremendous success of transformer in natural language processing (NLP), transformer~\cite{vaswani2017attention} has no significant influence on computer vision (CV) until the groundbreaking work by Dosovitskiy \etal~\cite{dosovitskiy2020image} proposes to split the image into patches and applies a pure transformer to process these patches like tokens in NLP. Their work shows competitive performance on image classification and reveals the great modeling capacity of transformer for vision tasks. The results galvanize researchers to bring ViTs into more vision tasks beyond classification. However, the quadratic computational cost of self-attention prevents ViTs from high-resolution input, which is common in visual recognition. To deal with this, Swin~\cite{liu2021swin} propose to restrict the self-attention in a fixed range by the local window, followed by other window-based approaches like cross-shape window~\cite{dong2022cswin}, pale-shape window~\cite{wu2022pale}, and~\cite{zhu2021long,yang2021focal}. Another line of works explores reducing the number of keys or values in self-attention via linear projection on reshaped spatial dimension~\cite{wang2021pyramid,chen2022regionvit}, strided convolution~\cite{wu2021cvt,zhang2022rest}, pooling~\cite{fan2021multiscale,si2022inception}, and clustering of patches~\cite{liu2022dynamic}. Other ViT variants apply various designs like channel attention~\cite{ali2021xcit,ding2022davit} (operating across feature channels instead of spatial dimension) and softmax-free attention~\cite{shen2021efficient,koohpayegani2022sima}.
\subsection{Efficient Attention}
How to address the quadratic computational cost of self-attention has attracted many researchers. Apart from the above efficient attentions in vision transformers, there are numerous methods in NLP~\cite{tay2020efficient}. They can be broadly categorized into the following categories: 1) sparse patterns~\cite{child2019generating,beltagy2020longformer,zaheer2020big,zhang2021poolingformer}, which sparsify the attention matrix using hand-crafted or learned pattern; 2) downsampling/low-rank~\cite{wang2020linformer,jaegle2021perceiver,xiong2021nystromformer}, which projects the key/value tensor into smaller tensor; 3) neural memory~\cite{sukhbaatar2019augmenting,lee2019set}, which leverages a side memory module for accessing multiple tokens; 4) linear attention, which decomposes the exponential kernel in softmax attention into dot product of kernel feature maps and is most related to our work. Katharopoulos~\etal~\cite{katharopoulos2020transformers} first propose linear attention and accelerate transformer in an iterative implementation like recurrent neural networks. Peng~\etal~\cite{peng2021random} use random feature methods to approximate the softmax function. Performer~\cite{choromanski2020rethinking} further introduces a positive orthogonal random feature mechanism. Moreover, cosFormer~\cite{qin2021cosformer} proposes a cosine-based distance re-weighting mechanism and achieves comparable accuracy. Most recently, Cai~\etal~\cite{cai2023efficientvit} first explore a light-weight linear attention with low computation. Han \etal~\cite{han2023flatten} propose an rank restoration module to enhance the expressiveness of self-attention.
In this paper, we investigate the reasons underlying the failure of linear attention in general-purpose vision transformers and design a novel concentration module to make linear attention competitive with vanilla attention.
\section{Preliminaries}
The attention mechanism is a core advantage of vision transformers over CNNs. Let $X\in \mathbb{R}^{N\times C}$ denote a sequence of N feature patches of dimension C, the vanilla softmax attention output $O\in \mathbb{R}^{N\times C}$ can be expressed as follows:
\begin{align} \label{equ:van attention}
	O_i=\sum_{j=1}^N{A_{ij} V_j}
	=\sum_{j=1}^N\frac{\text{exp}(Q_iK_j^T)}{\sum_{k=1}^N \text{exp}(Q_iK_k^T)}V_j.
\end{align}
The $A_i$ is the $i$th row in the learned attention matrix $A \in \mathbb{R}^{N\times N}$. $Q\in \mathbb{R}^{N\times C}$, $K\in \mathbb{R}^{N\times C}$, and $V\in \mathbb{R}^{N\times C}$ denotes query, key, and value matrix, generated by learnable linear projection $Q=W_QX$, $K=W_KX$, and $V=W_VX$, respectively. And $exp$ denotes the exponential function. Note that we omit the scale factor $1/\sqrt{C}$ for simplicity. 

The $exp$-based similarity is a specific form of similarity function, we can define a more generalized attention as:
\begin{align} \label{equ:sim attention}
	O_i=\sum_{j=1}^N\frac{\text{Sim}(Q_i, K_j^T)}{\sum_{k=1}^N \text{Sim}(Q_i, K_k^T)}V_j,
\end{align}
where Sim refers to the similarity function. Although softmax attention can build long-range dependency between N patches, it incurs a computation cost of $O(N^2C)$, infeasible for high-resolution input, \eg, N=66650 (1333/4$\times$800/4) after convolutional stem when the input size is 1333$\times$800. 
\paragraph{Linear Attention} To address this issue, linear attention~\cite{katharopoulos2020transformers} proposes to replace the exponential similarity function with decomposable kernel function $K$ as similarity function:
\begin{align}
	K(Q_i, K_j^T)=\phi(Q_i)\phi(K_j^T),
\end{align}
where $\phi$ refers to a random feature map. Thus the \cref{equ:sim attention} can be transformed as follows and further simplified using the associative property of matrix multiplication.
\begin{align} \label{equ:linear attention}
	\resizebox{0.9\hsize}{!}{
		$O_i=\sum_{j=1}^N\frac{\phi(Q_i)\phi(K_j^T)}{\sum_{k=1}^N \phi(Q_i)\phi(K_k^T)}V_j
		=\frac{\phi(Q_i)\sum_{j=1}^N\phi(K_j^T)V_j}{\sum_{k=1}^N \phi(Q_i)\phi(K_k^T)}.$
	}
\end{align}
The above equation reveals that linear attention can still capture dependency between all patches while reducing the computational cost from $O(N^2C)$ to $O(NC^2)$ by multiplying key and value first since query and key are decoupled, which makes linear attention especially attractive for downstream tasks including segmentation and detection where high-resolution feature maps are required.
\section{Method}

Some previous works~\cite{katharopoulos2020transformers, peng2021random, choromanski2020rethinking} have proposed different kernel function variants and achieved comparable results in NLP. Nevertheless, when researchers attempt to apply these linear attention mechanisms to vision transformer, the performance of linear variants lags far behind vanilla counterpart, \eg, 78.7\% (Performer) vs. 81.8\% (Vanilla)~\cite{hong2022fair}. These results suggest they all ignore some essential information for visual recognition. We re-examine linear attention from a visual perspective and show it can achieve on-par expressivity with softmax attention by incorporating two key properties.
\begin{table}[t]
	\begin{center}
		\setlength{\tabcolsep}{20pt}
		\resizebox{0.99\linewidth}{!}{
			\begin{tabular}{ccc}
				\hline
				Attention & $\phi$ &Top-1\\
				\hline
				SA.~\cite{touvron2021training}   & - & 72.2 \\
				SA.$^{\textbf{*}}$   & - &  72.5\\
				LA.  & L1 norm   & 68.6 \\
				LA.  & ReLU  &  69.3 \\
				LA.  &LeakyReLU   &  67.6 \\
				Enhanced LA. & ReLU & \textbf{73.3} \\
				\hline
			\end{tabular}
		}
	\end{center}
	\caption{\textbf{ImageNet-1k accuracy of DeiT-Tiny with different attention variants.} SA. is softmax attention, LA. is linear attention with feature map $\phi$. $^{*}$ indicates the results are reproduced by ours.}
	\label{tab:non-negative}
\end{table}
\begin{figure}[b]
	\begin{center}
		\includegraphics[width=0.5\textwidth]{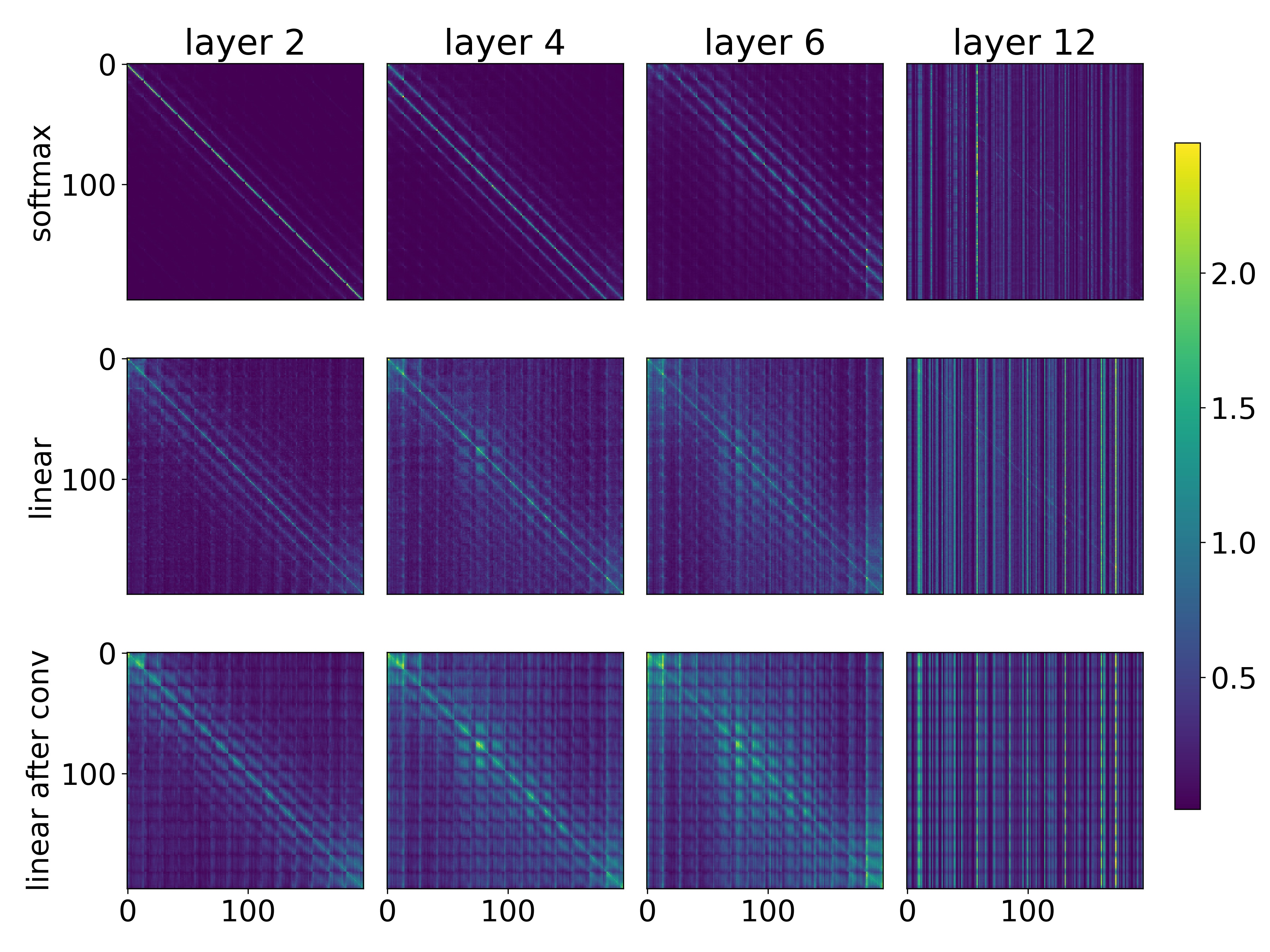}
	\end{center}
	\caption{\textbf{The attention maps of softmax, linear attention using ReLU as $\phi$, and local enhanced linear attention.} $x$ and $y$ axes indicate the patches. The deeper the network, the longer-range dependency the attention mechanism extracts.}
	\label{fig:non-negative}
\end{figure}

\begin{figure*}[t]
	\begin{center}
		\includegraphics[width=1.0\textwidth]{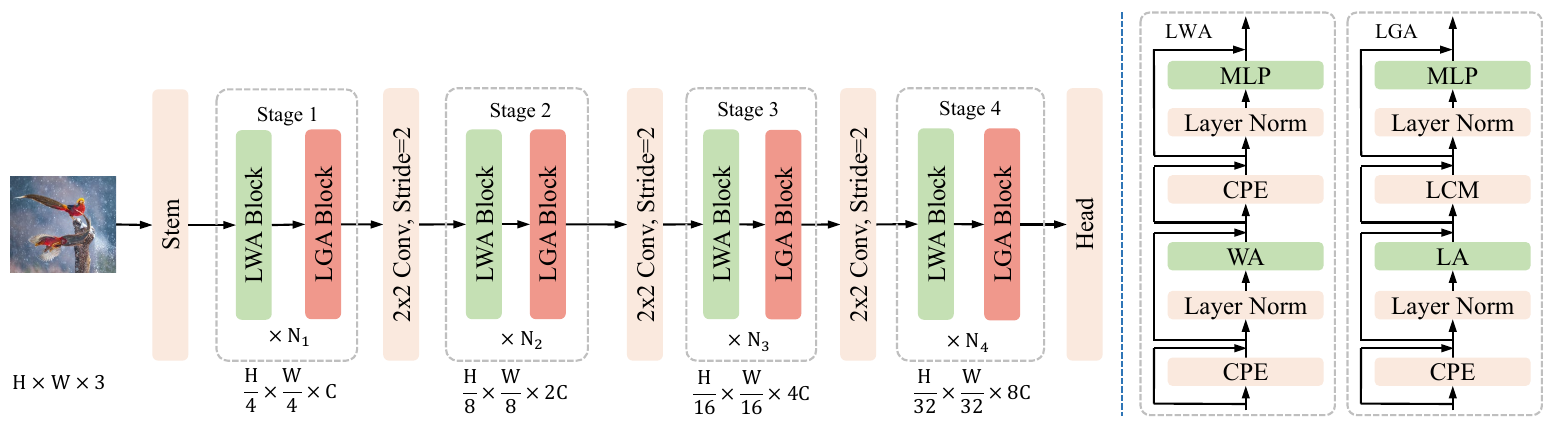}
	\end{center}
	\caption{Left: \textbf{the overall architecture of our proposed L$^2$ViT.} Right: \textbf{the illustration of the Local Window Attention block (LWA) and Linear Global Attention block (LGA)}. MLP indicates the \textbf{M}ulti-\textbf{L}ayer \textbf{P}erceptron. WA indicates Window Attention, and LA indicates Linear Attention.}
	\label{fig:overall}
	\vspace{-3mm}
\end{figure*}
\subsection{Non-negative Property}
First, the exponential function in vanilla attention forces all the values in the attention map $A$ to be non-negative. Although the value matrix $V$ contains negative values, unnecessary interactions (entries with values close to zero in $A$) still produce almost zero effect in the output. Instead, if the unnecessary interactions retain negative values in $A$, they may strengthen the irrelevant contextual information and disturb the attention contents. Inspired by~\cite{qin2021cosformer}, we replace the softmax attention in DeiT-Tiny~\cite{touvron2021training} with kernel-based linear attention of different feature map functions~$\phi$: 1) L1 norm~\cite{koohpayegani2022sima}, which keeps negative values; 2) ReLU, which guarantees the non-negative property; 3) LeakyReLU, which behaves similar to ReLU but allows negative values. We compare these designs in~\cref{tab:non-negative}. The stronger performance of ReLU over the L1 norm and LeakyReLU affirms the significance of the non-negative property.

Furthermore, different from sophisticated kernel functions in the previous works~\cite{katharopoulos2020transformers, peng2021random, choromanski2020rethinking}, ReLU is simple and efficient. By ensuring the non-negative property, ReLU-based linear attention is sufficient to extract short-range and long-range interactions as softmax attention. As shown in~\cref{fig:non-negative}, DeiT-Tiny with different attentions learns similar attention maps in both shallow and deep layers.

\subsection{Local Concentration Module}
Although linear attention can capture similar correlations as softmax attention, there is still a significant performance gap, as shown in~\cref{tab:non-negative}. We discover that the less favored performance of linear attention is mainly caused by the less concentrated attention map. Through re-weighting of softmax, vanilla attention can concentrate on important neighboring patches and other meaningful interactions as shown in~\cref{fig:non-negative}, \eg layer 4.
In contrast, linear attention presents a more dispersive map and trivially distributes attention scores over all patches. Although it can capture long-range dependencies, linear attention emphasizes neighboring patches less and preserves fewer local details as distracted by distant patches, potentially losing some essential fine-grained visual features of objects.

To further demystify these effects, we randomly pick some input images from ImageNet-1K~\cite{deng2009imagenet} and visualize the activation maps of DeiT-Tiny equipped with vanilla softmax and ReLU-based linear attention using Grad-CAM tool~\cite{selvaraju2017grad}. As clearly shown in~\cref{fig:grad-cam}, the former pays the most interest in the object itself, while the latter suffers from distractions of background and other stuff. These analyzes uncover that linear attention needs to concentrate more on important local information.

Motivated by the above observations, an intuitive way to preserve more local information is applying convolution following linear attention to distill the dispersive attention and reinforce local contextual features. Formally, recall that $O_j$ is the attention output for $j$-th patch in~\cref{equ:van attention}, the output enhanced by convolution can be written as:

\begin{align} \label{equ:enhanced attention}
	O^{'}_i=\sum_{j\in \Omega_i}w_jO_j=\sum_{k=1}^N{\sum_{j\in \Omega_i}w_jA_{jk} V_k},
\end{align}
where $\Omega_i$ is the local window centered at $i$ and $w_j$ is the convolution weight.  The above formulation explicitly shows that convolution can aggregate different rows in attention map $A$. \cref{fig:non-negative} provides visualization for the aggregated attention maps after convolution, in which neighboring patches (near the diagonal) receive stronger attention than that of linear attention without convolution. Specifically, we introduce a very lightweight local concentration module (LCM) consisting of two depth-wise convolutional layers:
\begin{align}
	& \hat{X} = \text{GELU}(\text{DWConv}_1(\text{Rearrange}(X)))\in \mathbb{R}^{C\times H\times W}, \\
	& X_{\text{LCM}} = \text{Rearrange}(\text{DWConv}_2(\text{BN}(\hat{X}))).
\end{align}
where $X$ is the output of the linear attention block. $H, W$ are the height and width of the feature map respectively. We can update these features by computing
\begin{align}
	Y=\text{LCM}(\text{LN}(X)) + X \in \mathbb{R}^{N\times C},
\end{align}
The details of LCM and our implementation are summarized in Appendix. We call linear attention followed by the LCM enhanced linear attention, illustrated in~\cref{fig:overall}. The results in~\cref{tab:non-negative} indicate that enhanced linear attention is more powerful. Meanwhile,~\cref{fig:grad-cam} also demonstrates that the proposed module helps linear attention focus on the object better.

Moreover, we find that directly applying~\cref{equ:linear attention} will cause unstable training and degrade performance due to large variance brought by multiplication. To counteract this effect, we clamp the denominator and apply a learnable scale parameter $s$ to scale down the dot-product of key and value. Thus we control the variance in a more stable range. More experimental results are presented in the Appendix.
\begin{table}[t!]
	\begin{center}
		\setlength{\tabcolsep}{12pt}
		\resizebox{0.85\linewidth}{!}{
			\begin{tabular}{l|ccc}
				\toprule
				Model & \makecell{\#Params. \\ (M)} & \makecell{FLOPs \\ (G)} & \makecell{Top-1 \\ (\%)} \\
				\midrule
				\multicolumn{4}{c}{ConvNets} \\  
				\midrule
				ConvNeXt-T~\cite{liu2022convnet} & 28 & 4.5 & 82.1 \\
				ConvNeXt-T~\cite{liu2022convnet} & 50 & 8.7 & 83.1 \\
				ConvNeXt-T~\cite{liu2022convnet} & 89 & 15.4 & 83.8 \\			
				EfficientNet-B4~\cite{tan2019efficientnet} & 19 & 4.2 & 82.9 \\
				EfficientNet-B5~\cite{tan2019efficientnet} & 30 & 9.9 & 83.6 \\
				EfficientNet-B6~\cite{tan2019efficientnet} & 43 & 19.0 & 84.0 \\
				\midrule
				\multicolumn{4}{c}{Vision Transformers} \\  
				\midrule		
				Swin-T~\cite{liu2021swin}   & 28 & 4.5 &  81.3 \\
				CoAtNet-0~\cite{dai2021coatnet} &25 & 4.2 &81.6 \\
				Twins-SVT-S~\cite{chu2021twins} & 24 & 2.9 & 81.7 \\
				SHViT-S4~\cite{yun2024shvit} & 16.5 & 4.0 & 82.0 \\
			    FasterViT-0~\cite{hatamizadeh2023fastervit} &31 & 3.3 & 82.1 \\
				Flatten-Swin-T~\cite{han2023flatten} & 29 & 4.5 & 82.1 \\
				Focal-Tiny~\cite{yang2021focal} & 29 &4.9 & 82.2 \\
				ResTv2-T~\cite{zhang2022rest} & 30 &4.1 & 82.3 \\
				RepViT-M2.3~\cite{wang2024repvit} & 23 & 4.5 & 82.5 \\
				CrossFormer-S~\cite{wang2021crossformer} &31 &4.9 &82.5 \\
				Agent-Swin-T~\cite{han2023agent} & 29 & 4.5 & 82.6 \\
				EfficientViT-B2\cite{cai2023efficientvit} &24 &- &82.7 \\
				CETNet-T~\cite{wang2022convolutional} & 23 & 4.3 & 82.7 \\
				DaViT-T$^{*}$~\cite{ding2022davit} & 28 & 4.5 & 82.7 \\
				MPViT-S~\cite{lee2022mpvit} & 23 & 4.7 &83.0 \\
				L$^2$ViT-T~(ours)              & 29 & 4.7 &  \textbf{83.1} \\		
				\midrule    
				Swin-S~\cite{liu2021swin}   & 50 & 8.7 &  83.2 \\
				Twins-SVT-B~\cite{chu2021twins} & 56 & 8.6 & 83.2 \\
				CoAtNet-1~\cite{dai2021coatnet} &42 &8.4 & 83.3 \\
				Focal-Small~\cite{yang2021focal} & 51 &9.4 &83.6 \\
				CrossFormer-B~\cite{wang2021crossformer} &52 & 9.2 &83.4 \\
				RegionViT-M+~\cite{chen2022regionvit} &42 & 7.9 & 83.4 \\
				CETNet-S~\cite{wang2022convolutional} & 34 & 6.8 &83.4 \\
				EfficientViT-B3\cite{cai2023efficientvit} &49 &- &83.5 \\
				Flatten-Swin-S~\cite{han2023flatten} & 51 & 8.7 & 83.5 \\
				ResTv2-B~\cite{zhang2022rest} & 56 & 7.9 & 83.7 \\
				Agent-Swin-T~\cite{han2023agent} & 50 & 8.7 & 83.7 \\
				DaViT-S$^{*}$~\cite{ding2022davit} & 50 & 8.8 & 83.8 \\
				XCiT-S24/16$^{\dagger}$~\cite{ali2021xcit} & 48 & 9.1 & 83.9 \\
				L$^2$ViT-S~(ours)              & 50 & 9.0 &  \textbf{84.1} \\	
				\midrule
				Swin-B~\cite{liu2021swin}   & 88 & 15.4 &  83.5 \\
				Twins-SVT-L~\cite{chu2021twins} & 99 & 15.1 & 83.7 \\
				RegionViT-B+~\cite{chen2022regionvit} & 74 &13.6 &83.8 \\
				CETNet-B~\cite{wang2022convolutional} & 75 & 15.1 &83.8 \\
				Flatten-Swin-B~\cite{han2023flatten} & 88 & 15.4 & 83.8 \\
				DaViT-B$^{*}$~\cite{ding2022davit} & 88 & 15.5 & 83.9 \\
				Focal-Base~\cite{yang2021focal} & 90 & 16.4 & 84.0 \\
				CrossFormer-L~\cite{wang2021crossformer} & 92 & 16.1 &84.0 \\
				Agent-Swin-T~\cite{han2023agent} & 88 & 15.4 & 84.0 \\
				CoAtNet-2~\cite{dai2021coatnet} &75 &15.7 & 84.1 \\
				ResTv2-L~\cite{zhang2022rest} &87 & 13.8 & 84.2 \\
				MPViT-B~\cite{lee2022mpvit} & 75 & 16.4 & 84.3 \\
				XCiT-M24/16$^{\dagger}$~\cite{ali2021xcit} & 84 & 16.2 & 84.3 \\
				L$^2$ViT-B~(ours)   & 89 & 15.9 &  \textbf{84.4} \\	
				\bottomrule
			\end{tabular}
		}
	\end{center}
	\caption{\textbf{Classification performance on ImageNet-1K.} All models are trained with 224 $\times$ 224 resolution except EfficientNet~\cite{tan2019efficientnet}. $\dagger$ indicates models are trained with distillation. $^{*}$ indicates that we use the official public implementation and reproduce the results using a cosine learning rate schedule for fair comparison as the original paper~\cite{ding2022davit} uses a triangular schedule.}
	\label{tab:cls_1k}
	\vspace{-2pt}
\end{table}
\subsection{Overall Architecture}
We integrate the local concentration module (LCM) and linear attention to build a Linear Global Attention block (LGA), which captures the global contextual information. Meanwhile, we employ window attention~\cite{liu2021swin} to build a Local Window Attention block (LWA), which introduces an ideal locality and refines the fine-grained feature representations. These two complementary blocks are stacked alternatively to design an efficient and general-purpose vision transformer dubbed L$^2$ViT. 

The overall architecture and block details are illustrated in~\cref{fig:overall}. We employ a hierarchical framework to obtain pyramid feature maps for a broad range of visual recognition tasks. Given an input image with size $H\times W\times 3$, we leverage a convolutional stem (two 3$\times$3 convolutional layers with stride 2) to obtain $\frac{H}{4}\times \frac{W}{4}$ patches with dimension $C$. Then all patches go through the following four stages, each stage $i\in \text{(1,2,3,4)}$ contains $N_i$ LWA and $N_i$ LGA blocks alternatively. Between stages, we use another convolutional layer (2$\times$ 2, stride 2) to merge patches and double the dimension. Especially, we introduce the flexible Conditional Positional Encodings (CPE)~\cite{chu2021conditional} to replace the relative position embedding in every block.

We build several L$^2$ViT variants with different FLOPs and number of parameters. The detailed configuration is provided in Appendix~C. In all variants, for a fair comparison with previous works, we keep the strictly same number of blocks, heads, and channels as Swin~\cite{liu2021swin}, while deepening the depth will improve the performance as shown in~\cref{tab:abl_aug}.

\section{Experiments}
\subsection{ImageNet-1K Classification}
\begin{table}[t!]
	\begin{center}
		\setlength{\tabcolsep}{12pt}
		\resizebox{0.95\linewidth}{!}{
			\begin{tabular}{l|cccc}
				\toprule
				Model & \makecell{Image \\size} & \makecell{\#Params. \\ (M)} & \makecell{FLOPs \\ (G)} & \makecell{Top-1 \\ (\%)} \\
				\midrule
				Swin-B~\cite{liu2021swin}   & 224 & 88 & 15.4 &  85.2 \\
				Swin-B~\cite{liu2021swin}   & 384 & 88 & 47.0 &  86.4 \\
				\midrule
				RegionViT-B+~\cite{chen2022regionvit}  & 384 & 77 & 42.6 &  86.5 \\
				\midrule
				L$^2$ViT-B~(ours)  & 224  & 89 & 15.9 &  \textbf{86.0} \\		
				L$^2$ViT-B~(ours)  & 384  & 89 & 47.5 &  \textbf{87.0} \\	
				\bottomrule
			\end{tabular}
		}
	\end{center}
	\caption{\textbf{ImageNet-1k fine-tune results with pre-training on ImaegNet-22k.}}
	\label{tab:cls_22k}
	\vspace{-6pt}
\end{table}
\begin{table*}[ht]
	\begin{center}
		\renewcommand\arraystretch{.95}
		\resizebox{0.99\linewidth}{!}{
			\setlength{\tabcolsep}{5.5pt}{
				\begin{tabular}{l@{\hspace{3.2pt}}|c@{\hspace{3.2pt}}|c|c|c|c|c|c|c|c|c|c|c|c|c}
					\toprule
					\multirow{2}{*}{Backbone} & \#Params & FLOPs & \multicolumn{6}{c|}{Retina 1x schedule} & \multicolumn{6}{c}{Mask R-CNN 1x schedule}\\
					& (M) & (G) & $AP^b$ & $AP^b_{50}$ & $AP^b_{75}$ & $AP^b_{S}$ & $AP^b_{M}$ & $AP^b_{L}$ & $AP^b$ & $AP^b_{50}$ & $AP^b_{75}$ & $AP^m$ & $AP^m_{50}$ & $AP^m_{75}$ \\
					\midrule
					ResNet50~\cite{he2016deep}     & 38/44 & 239/260  & 36.3 & 55.3 & 38.6 & 19.3 & 40.0 & 48.8
					& 38.0 & 58.6 & 41.4 & 34.4 & 55.1 & 36.7 \\
					PVT-S~\cite{wang2021pyramid}   & 34/44 & 226/245  & 40.4 & 61.3 & 43.0 & 25.0 & 42.9 & 55.7
					& 40.4 & 62.9 & 43.8 & 37.8 & 60.1 & 40.3 \\
					Swin-T~\cite{liu2021swin}   & 39/48 & 245/264  & 42.0 & 63.0 & 44.7 & 26.6 & 45.8 & 55.7
					& 43.7 & 66.6 & 47.7 & 39.8 & 63.3 & 42.7 \\
					Twins-SVT-S~\cite{chu2021twins}   & 34/44 & 216/245  & 43.0 & 64.2 & 46.3 & 28.0 & 46.4 & 57.5
					& 43.4 & 66.0 & 47.3 & 40.3 & 63.2 & 43.4 \\
					FLatten-Swin-T~\cite{han2023flatten} & -/49 & -/268 & - & - & - & - & - & - & 44.2 & 67.3 & 48.5 & 40.2 & 63.8 & 43.0 \\
					RegionViT-S+w/PEG~\cite{chen2022regionvit} & 42/51 & 204/183 & 43.9 &65.5 & 47.3 & 28.5 & 47.3 & 57.9 & 44.2 & 67.3 & 48.2 & 40.8 & 64.1 & 44.0 \\
					CMT-S~\cite{guo2022cmt} & 35/45 & 231/249 & \textbf{44.3} & 65.5 & \textbf{47.5} & 27.1 & \textbf{48.3} & \textbf{59.1} &
					44.6 & 66.8 & 48.9 & 40.7 & 63.9 & 43.4 \\
					CrossFormer-S~\cite{wang2021crossformer} & 41/50 & 272/291 & 44.2 & 65.7 & 47.2 & 28.0 & 48.0 & 59.1
					& 45.0 & 67.9 & 49.1 & 41.2 & 64.6 & \textbf{44.3} \\
					CETNet-T~\cite{wang2022convolutional} & -/43 & -/261 & - & - &- &- &- & - 
					& 45.5 & 67.7 & 50.0 & 40.7 & 64.4 & 43.7 \\
					L$^2$ViT-T(ours)            & 39/48 & 250/269  & 44.1 & \textbf{65.8} & 47.0 & \textbf{30.0} & 47.9 & 57.9
					& \textbf{45.5} & \textbf{68.6} & \textbf{50.1} & \textbf{41.2} & \textbf{65.1} & 44.0\\
					
					\midrule
					ResNeXt101-32x4d~\cite{xie2017aggregated}   & 56/63 & 319/340  & 39.9 & 59.6 & 42.7 & 22.3 & 44.2 & 52.5 
					& 41.9 & 62.5 & 45.9 & 37.5 & 59.4 & 40.2\\
					PVT-M~\cite{wang2021pyramid}           & 54/64 & 283/302  & 41.9 & 63.1 &44.3 & 25.0 & 44.9 & 57.6
					& 42.0& 64.4 & 45.6 & 39.0 & 61.6 & 42.1 \\
					Swin-S~\cite{liu2021swin}   & 60/69 & 335/354  & 44.5 & 65.7 & 47.5 & 27.4 & 48.0 & 59.9
					& 44.8 & 66.6 & 48.9 & 40.9 & 63.4 & 44.2 \\
					Twins-SVT-B~\cite{chu2021twins}   & 67/76 & 337/357  & 45.3 & 66.7 & 48.1 &  28.5 & 48.9 & 60.6
					& 45.2 & 67.6 & 49.3 & 41.5 & 64.5 & 44.8 \\
					RegionViT-B+w/PEG~\cite{chen2022regionvit} & 85/93 & 328/307 & 44.6 &66.4 & 47.6 & 29.6 & 47.6 & 59.0 & 45.4 & 68.4 & 49.6 & 41.6 & 65.2 & 44.8 \\
					CrossFormer-B~\cite{wang2021crossformer} &62/72 & 379/398 & 46.1 & 67.7 & 49.0 & 29.5 & 49.9& \textbf{61.5}  
					& 47.1 & 69.9 & \textbf{52.0} & \textbf{42.7} & \textbf{66.5} & \textbf{46.1} \\
					CETNet-S~\cite{wang2022convolutional} & -/53 & -/315 & - & - &- &- &- & - 
					& 46.6 & 68.7 & 51.4 & 41.6 & 65.4 & 44.8 \\
					L$^2$ViT-S(ours)      & 60/70 & 341/360  & \textbf{46.2} & \textbf{68.0} & \textbf{49.6} & \textbf{31.7} & \textbf{50.0} & 61.1
					& \textbf{47.2} & \textbf{70.0} & 51.7 & 42.4 & 66.4 & 45.2 \\
					
					\midrule
					ResNeXt101-64x4d~\cite{xie2017aggregated}   & 96/102 & 473/493  & 41.0 & 60.9 & 44.0 & 23.9 & 45.2 & 54.0
					& 42.8 & 63.8 & 47.3 & 38.4 & 60.6 & 41.3 \\
					PVT-L~\cite{wang2021pyramid}    & 71/81  & 345/364  & 42.6 & 63.7 & 45.4 & 25.8 & 46.0 & 58.4
					& 42.9 & 65.0 & 46.6 & 39.5 & 61.9 & 42.5 \\
					Swin-B~\cite{liu2021swin}   & 98/107 & 477/496  & 44.7 & - & - & - & - & -
					& 45.5 & - & - & 41.3 & - & - \\
					Twins-SVT-L~\cite{chu2021twins}   & 111/120 & 455/474  & 45.7 & - & - &  - & - & -
					& 45.9 & - & - & 41.6 & - & - \\
					RegionViT-B+w/PEG$^{\dagger}$~\cite{chen2022regionvit} & 85/93 & 506/464 & 46.1 &68.0 & \textbf{49.5} & 30.5 & 49.9 & 60.1 & 46.3& 69.1 & 51.2 & 42.4 & 66.2 & 45.6 \\
					CETNet-B~\cite{wang2022convolutional} & -/94 & -/495 & - & - &- &- &- & - 
					& \textbf{47.9} & 70.3 & \textbf{53.0} & 42.5 & \textbf{67.2} & 45.6 \\
					L$^2$ViT-B(ours)     & 99/108  & 484/504  & \textbf{46.5} & \textbf{68.7} & 49.4 & \textbf{31.2} & \textbf{50.6} & \textbf{60.7}
					& 47.5 & \textbf{70.5} & 51.8 & \textbf{42.9} & 67.1 & \textbf{45.9} \\
					\bottomrule
				\end{tabular}
		}}
	\end{center}
	\caption{\textbf{Object detection and instance segmentation performance on COCO with Retina and Mask R-CNN framework}. The FLOPs are measured at resolution 800 $\times$ 1280. All models are pre-trained on the ImageNet-1k and fine-tuned on the COCO 2017 using 1x training schedule. $\dagger$ indicates input resolution is 896$\times$ 1344.}
	\label{tab:det_1x}
\end{table*}
For a fair comparison, we train our models for 300 epochs following the recipe in~\cite{touvron2021training, liu2021swin,liu2022convnet}. More details are provided in Appendix.~\cref{tab:cls_1k} compares our L$^2$ViT with state-of-the-art ConvNets and Vision Transformers trained only on ImageNet-1k. L$^2$ViT achieves stronger performance under different model sizes and computational complexities. Compared to ConvNets, our L$^2$ViT has better accuracy. Especially, while most vision transformers show unsatisfactory results on the small size compared to EfficientNet, L$^2$ViT-T obtains an improved result of 83.1\%.

Our L$^2$ViT outperforms other vision transformers, including Swin (fixed-size window attention) and Twins-SVT (mixing window and keys/values reduction attention). For example, L$^2$ViT-B achieves an accuracy of 84.4\%, surpassing Swin-B and Twin-SVT-L by +0.9\% and +0.7\%, respectively. This shows the superiority of enhanced linear attention in capturing the global context. Meanwhile, L$^2$ViT outperforms channel attention-based DaViT, by a large margin. For example, L$^2$ViT-B achieves +0.5\% higher accuracy than DaViT-B, indicating that global patch-to-patch interactions play a more critical role than channel-to-channel interactions.

Besides, we pre-train L$^2$ViT-B on the larger scale ImageNet-22k using 224$^2$ and 384$^2$ input size.~\cref{tab:cls_22k} shows ImageNet-22k data adds +1.6\% accuracy and larger input size adds +1.0\% accuracy to L$^2$ViT-B. L$^2$ViT-B with 384$^2$ input size achieves 87.0\% accuracy, surpassing prior models. The pre-training results further demonstrate the strong model capacity of L$^2$ViT.

\subsection{COCO Object Detection}
We conduct object detection experiments on COCO dataset~\cite{lin2014microsoft} using standard Mask R-CNN~\cite{he2017mask} and Retina~\cite{lin2017focal} detection framework implemented in MMdetection Toolboxes~\cite{mmdetection}. 
For a fair comparison, we follow the same recipe as Swin~\cite{liu2021swin}. 

\cref{tab:det_1x} summarizes the results measured by both box and mask mAP. 
For detection results with Retina, L$^2$ViT outperforms Swin and Twins-SVT by a large margin. For example, L$^2$ViT-T improves over Swin-T and Twins-SVT-S by +2.1 and +1.1 AP$^{b}$ respectively. This further shows that enhanced linear attention indeed extracts a richer representation and enables the model to detect objects better. 

For detection results with Mask R-CNN, L$^2$ViT brings clear improvements over Swin and Twins-SVT in different model sizes. Meanwhile, L$^2$ViT-S surpasses CETNet-S +0.6/+0.8 in AP$^{b}$
/AP$^{m}$. Although L$^2$ViT-B performs slightly worse than CETNet-B in AP$^{b}$, the results on AP$^{m}$ are reversed. We also stress that CETNet applies a deep-narrow model ([4,4,30,2] of CETNet-B vs. [2,2,18,2] of ours) that brings extra gains as shown in~\cref{tab:abl_aug}. The improved performance on both detection frameworks validates the generalizability of our proposed L$^2$ViT.
\subsection{ADE20K Semantic Segmentation}
\begin{table}[t]
	\begin{center}
		\setlength{\tabcolsep}{12pt}
		\resizebox{0.9\linewidth}{!}{
			\begin{tabular}{l|cccc}
				\toprule
				Backbone & Crop Size & \#Param.(M) &FLOPs(G)& mIoU \\
				\midrule
				Swin-T~\cite{liu2021swin} & 512$\times$ 512 & 60 & 945 & 44.5  \\
				XCiT-S12/16~\cite{ali2021xcit} & 512$\times$ 512 & 52 & - & 45.9 \\
				Twins-SVT-S~\cite{chu2021twins} & 512$\times$ 512 & 54 & 912 & 46.2 \\
				Focal-T~\cite{yang2021focal} & 512 $\times$ 512 & 62 & 998 & 45.8 \\ 
				L$^2$ViT-T(ours) & 512$\times$ 512 & 60 & 943 & \textbf{46.2} \\   
				\midrule 
				Swin-S~\cite{liu2021swin} & 512$\times$ 512 & 81 & 1038 & 47.6 \\  
				XCiT-S24/16~\cite{ali2021xcit} & 512$\times$ 512 & 74 & - & 46.9 \\
				Twins-SVT-B~\cite{chu2021twins} & 512$\times$ 512 & 89 & 1044 & 47.4 \\
				Focal-S~\cite{yang2021focal} & 512 $\times$ 512 & 85 & 1130 & 48.0 \\		 
				L$^2$ViT-S(ours) & 512$\times$ 512 & 82 & 1034 & \textbf{48.7}  \\  
				\midrule     
				Swin-B~\cite{liu2021swin} & 512$\times$ 512 & 121 & 1188 & 48.1 \\   
				XCiT-M24/16~\cite{ali2021xcit} & 512$\times$ 512 & 109 & - & 47.6 \\
				Twins-SVT-L~\cite{chu2021twins} & 512$\times$ 512 & 133 & 1188 & 48.8 \\ 
				Focal-B~\cite{yang2021focal} & 512 $\times$ 512 & 126 & 1354 & 49.0 \\	 
				L$^2$ViT-B(ours) & 512$\times$ 512 & 122 & 1182 & \textbf{49.2} \\      
				\bottomrule
			\end{tabular} 
		}
	\end{center}
	\caption{\textbf{Semantic segmentation performance on ADE20K~\cite{zhou2017scene}}. The FLOPs are measured at resolution 512 $\times$ 2048. 
	}
	\label{tab:seg}
	\vspace{-2pt}
\end{table}
We conduct semantic segmentation experiments on ADE20K~\cite{zhou2017scene} dataset. We adopt the UperNet~\cite{xiao2018unified} segmentation framework implemented in MMSegmentation Toolboxes~\cite{mmseg2020}. Following Swin~\cite{liu2021swin}, we train all models for 160k iterations with a batch size of 16, AdamW optimizer, multi-scale training, and stochastic depth. 

We present the result in~\cref{tab:seg}. Consistent improvements over Swin and Twins-SVT can be observed. In detail, L$^2$ViT-S achieves +1.1 and +1.3 higher mIOU than Swin-S and
 Twins-SVT-B. L$^2$ViT also outperforms other vision transformers under all model sizes, \eg, L$^2$ViT-T/S/B exceeds Focal-T/S/B by +0.4, +0.7, and +0.2 mIOU, respectively. The superior performance on semantic segmentation further demonstrates the effectiveness of enhanced linear attention and expressivity of L$^2$ViT.
\subsection{Ablation Study} \label{exp:abl}
We train all models for 300 epochs on ImageNet-1k and fine-tune Mask R-CNN for 1x schedule.
\paragraph{Component Analysis}
To study the effectiveness of key components in L$^2$ViT, we make several architecture changes and report the results in~\cref{tab:abl_component}. It can be observed that: 1) shrinking the kernel size in LCM into 3$\times$3 causes a dramatic drop in classification, which indicates that a large receptive field of LCM is important for concentrating interactions; 2) without LCM, L$^2$ViT will degenerate heavily both on classification and object detection, this reveals that concentrating the attention map locally contributes to better recognization, especially on dense prediction tasks; 3) scale parameter has a slight effect on the performance, but it improves the training stability; and 4) we further ablate convolutional stem and apply the patchify stem as Swin, which we call primitive L$^2$ViT. Meanwhile, we construct a new model named Enhanced Swin-T-V1 by replacing the relative position embedding with CPE for a fair comparison. 
Obviously, primitive L$^2$ViT yields slightly better accuracy than Enhanced Swin-T-V1 (+0.1\%), suggesting that models utilizing linear attention can outperform those employing local window attention, even in the absence of a LCM. 
To further show the effectiveness of the enhanced global linear attention, we directly replace linear attention in L$^2$ViT with window attention, resulting a model we refer to as Enhanced Swin-T-V2. The results clearly indicates that it still lags behind L$^2$ViT by 0.6\% in image classification and 0.7 AP$^b$ in object detection. This finding underscores the significance of incorporating linear attention in conjunction with the LCM, as opposed to relying solely on the LCM for performance enhancement. Compared to Swin-T-V1 , Swin-T-V2 exhibits a slight imrovement. It is important to note that the LCM possesses a larger receptive field than CPE, yet the 7x7 receptive field remains equivalent to that of window attention and is smaller than that of linear attention. These results suggest that the proposed enhanced linear atttention can capture more global and comprehensive representations than local window attention.
\begin{table}[t]
	\begin{center}
		\setlength{\tabcolsep}{6pt}
		\footnotesize
		\resizebox{0.95\linewidth}{!}{
			\begin{tabular}{l|cccccccc}
				\toprule
				& CPE &Conv & Scale & LCM &  \#Params.(M)/  & Top-1& COCO \\
				& &Stem   & Parameter &  & FLOPs(G) & (\%) & AP$^b$\\
				\toprule
				Enhanced Swin-T-V1 & \cmark& \xmark & \xmark & \xmark & 28/4.5 & 82.3 & 44.7 \\
				\midrule
				Enhanced Swin-T-V2& \cmark & \cmark & \xmark & \cmark & 29/4.7 & 82.5 & 44.8 \\
				\midrule
				\multirow{5}{*}{L$^2$ViT-T} & \cmark & \xmark& \xmark& \xmark & 28/4.5 & 82.4 & 44.8 \\
				& \cmark& \cmark& \xmark& \xmark & 28/4.5 & 82.4 & 44.8 \\
				& \cmark& \cmark& \cmark& \xmark & 28/4.5 & 82.5 & 44.7 \\
				& \cmark& \cmark& \cmark& kernel 3x3 & 29/4.6 & 82.7 & 45.3 \\
				& \cmark& \cmark& \cmark& kernel 7x7 & 29/4.7 & 83.1 & 45.5 \\
				\bottomrule
			\end{tabular}
		}
	\end{center}
	\caption{\textbf{Component analysis for L$^2$ViT.} The LCM kernel is 7$\times$7 in L$^2$ViT-T.}
	\label{tab:abl_component}
	\vspace{-2pt}
\end{table}

\paragraph{Model Augmentaion}
\begin{table}[h]
	\begin{center}
		\setlength{\tabcolsep}{2pt}
		\resizebox{0.95\linewidth}{!}{
			\footnotesize
			\centering
			\begin{tabular}{l|ccccc}
				\toprule
				&  \#Params. & FLOPs & ImaegNet 1k & \multicolumn{2}{c}{COCO} \\
				& (M) & (G) & Top-1 (\%) & AP$^b$ & $AP^m$\\
				\toprule
				L$^2$ViT-T & 29 & 4.7 & 83.1 & 45.5 & 41.2 \\
				\midrule
				\makecell[l]{+ Deep-narrow arch.~\cite{wang2022convolutional} \\ Depth:[2,2,6,2] $\rightarrow$ [4,4,18,4]} &
				25 & 4.4 & 83.2 & 46.1 & 41.6\\
				\midrule
				+ 4L conv stem~\cite{wang2022scaled} & 25 & 4.6 & 83.2 & 45.7 & 41.1 \\
				\midrule
				\makecell[l]{+ Projection layer before head~\cite{guo2022cmt} \\ 512 $\times$ 1280 }
				& 26 & 4.6 & 83.4 & 46.0 &41.4 \\
				\midrule
				+ Overlapped downsample layer~\cite{wu2021cvt}
				& 27 & 4.7 & 83.5 & 46.3 & 41.6 \\	     	
				\bottomrule
			\end{tabular}
		}
	\end{center}
	\caption{\textbf{Apply other model augmentation techniques to L$^2$ViT sequentially.} All variants share similar computational complexity.}
	\label{tab:abl_aug}
	\vspace{-6pt}
\end{table}
Above, we conduct a strictly fair comparison with previous models. Recent vision transformers~\cite{wu2021cvt,wang2022convolutional,wu2022pale,wang2022scaled,guo2022cmt, yu2022metaformer} introduce some orthogonal techniques such as deep-narrow architecture layout to obtain better performances. Here we investigate whether these common techniques are able to improve L$^2$ViT in~\cref{tab:abl_aug}. First, we design a deep-narrow variant where the base channel dimension reduces from 96 to 64. We see that deep-narrow layout brings a significant gain on object detection (+0.6 AP$^{b}$ and +0.4 AP$^{m}$) and a slight +0.1\% gain of ImageNet-1k accuracy. Second, as most works show that convolutions in shallow layers contribute much to ViT, we continue to add a 4-layer convolutional stem. However, it degrades the performance of detection. 
Third, we continue to add an extra projection layer with 1280 channels before the classification head to preserve more details. Although the projection layer is not added in the detection backbone, it still improves detection performance, suggesting that it can provide better initialization. On top of the aforementioned augmentations, we inject more inductive bias into L$^2$ViT by enlarging the convolutional kernel in downsampling layers from 2$\times$2 to 3$\times$3, which lifts both classification and detection performance. All these changes lead to clear improvements on various tasks. Furthermore, other techniques, such as Inverted Residual Feed-forward Network~\cite{guo2022cmt} may also boost the performance of L$^2$ViT.

\paragraph{More Attention Variants}
\begin{table}[t]
	\begin{center}
		\setlength{\tabcolsep}{6pt}
		\footnotesize
		\resizebox{0.92\linewidth}{!}{
			\begin{tabular}{c|ccc}
				\toprule
				Attention Variants & \# Params.(M) & FLOPs(G) & Top-1 (\%) \\
				\midrule
				Softmax Attention~\cite{touvron2021training} & 5.7 & 1.3 & 72.2 \\
				XCA~\cite{ali2021xcit} & 5.7 & 1.1 & 68.1 \\
				cosFormer~\cite{qin2021cosformer} & 5.7 & 1.4 & 67.7 \\
				EfficientAttention~\cite{shen2021efficient} & 5.7 & 1.3 & 67.7 \\
				SimA~\cite{koohpayegani2022sima} & 5.7 & 1.3 & 68.6 \\
				Linear Attention & 5.7 & 1.3 & 69.3 \\
				Enhanced Linear Attention & 5.8 & 1.3 & \textbf{73.3} \\
				\bottomrule
			\end{tabular}
		}
	\end{center}
	\caption{\textbf{Comparison of different attention variants.} We replace the original softmax attention in DeiT-T with different attention mechanisms.}
	\label{tab:abl_variant}
	\vspace{-6pt}
\end{table}
\cref{tab:abl_variant} compares enhanced linear attention with different attention mechanisms. To avoid influence caused by other factors such as CPE, we conduct ablation on DeiT-Tiny and replace all vanilla softmax attentions with channel attention (XCA~\cite{ali2021xcit}), linear attention with cos-based re-weighting mechanism (cosFormer~\cite{qin2021cosformer}), dot-product attention $\phi(Q)\phi(K)V$ using softmax function~\cite{shen2021efficient} or $\mathcal{\ell}_{1}$ norm~\cite{koohpayegani2022sima} as $\phi$, linear attention, and enhanced linear attention. Obviously, linear attention outperforms other attention variants. Although channel attention also captures the global receptive field, it performs inferiorly because it ignores patch-to-patch interactions. We also notice that the cos-based re-weighting mechanism is unsuitable for visual recognition. However, there is still a big gap between softmax attention and linear attention. By integrating the proposed LCM, our enhanced linear attention can focus on more neighboring interactions and achieve exciting accuracy. More ablation experiments and limitations are discussed in Appendix.

\section{Conclusion}
We present a new general-purpose vision transformer named L$^2$ViT, composed of two effective self-attention mechanisms (LGA and LWA). The appealing LGA develops a highly effective enhanced linear attention to build global long-range contextual relationships in linear complexity. At the same time, LWA employs well-designed window attention to focus on fine-grained local information. Taken these representations together, L$^2$ViT can better model the nature of our visual world and shows strong performance on various tasks,
suggesting strong potential for widespread applications.

{\small
\bibliographystyle{ieee_fullname}
\bibliography{l2vit}
}

\clearpage
\section*{Appendix}
\section{Training Details} \label{app:training detail}
We follow the training strategy in \cite{liu2021swin, liu2022convnet} and show the setting in~\cref{tab:train_detail}. When fine-tuning the 22k pre-trained model on ImageNet-1k, we use the same fine-tuning strategy as Swin~\cite{liu2021swin}. Specifically, we fine-tune the models for 30 epochs with a batch size of 1024, an initial learning rate of 1e-04, 5 epochs of linear warm-up, and a stochastic drop rate of 0.2. Particularly, we also apply 12$\times$12 window size in LWA when fine-tuning on ImageNet-1k with 384$\times$ 384 input. Furthermore, in all models, we clamp the denominator of Eq. (4) into the range $[1e2, +\infty)$. The learnable scale parameter $s$ is initialized as $\sqrt{C}$. 
\begin{table}[h]
	\begin{center}
		\setlength{\tabcolsep}{15pt}
		\footnotesize	
		\resizebox{0.99\linewidth}{!}{
			\begin{tabular}{l|cc}
				\multirow{2}{*}{training config} & L$^2$ViT-T/S/B & L$^2$ViT-B\\
				& ImageNet-1K & ImageNet-22K \\
				\toprule
				optimizer & AdamW~\cite{loshchilov2017decoupled} & AdamW\\
				batch size & 4096 & 4096 \\
				training epochs & 300 & 90 \\
				base learning rate & 4e-3 & 1e-3 \\
				weight decay & 0.05 & 0.05 \\
				learning rate schedule & \makecell{cosine decay\\by step} & \makecell{cosine decay\\by step} \\
				warmup epochs & 20 & 5 \\
				warmup schedule & linear & linear \\
				randaugment \cite{cubuk2020randaugment} & (9, 0.5) & (9, 0.5) \\
				mixup \cite{zhang2017mixup} & 0.8 & 0.8 \\
				cutmix \cite{yun2019cutmix} & 1.0 & 1.0 \\
				random erasing \cite{zhong2020random} & 0.25 & 0.25 \\
				label smoothing \cite{szegedy2016rethinking} & 0.1 & 0.1 \\
				stochastic depth \cite{huang2016deep} & 0.1/0.3/0.4 & 0.2 \\
				gradient clip & None & None \\
				EMA \cite{polyak1992acceleration} & 0.9999 & None\\
		\end{tabular}}
	\end{center}
	\caption{\textbf{ImageNet-1K training and 22K pre-training settings.}}
	\label{tab:train_detail}
\end{table}

\section{Implement Details for Clamping}
\begin{table}[h]
	\begin{center}
		\setlength{\tabcolsep}{6pt}
		\footnotesize
		\resizebox{0.99\linewidth}{!}{
			\begin{tabular}{c|ccc}
				\toprule
				\multirow{2}{*}{Lower-bound $C_{min}$ of Clamp } & ImaegNet 1k & \multicolumn{2}{c}{COCO} \\
				& Top-1 (\%) &  AP$^b$ & $AP^m$\\
				\midrule
				1e-6 & fail & fail & fail \\
				1e-1 & 82.7 & 44.6 & 40.6 \\
				1e0  & 82.9 & 45.0 & 41.0 \\
				1e1  & 82.9 & 45.1 & 41.0 \\
				1e2 & \textbf{83.1} & \textbf{45.5} & \textbf{41.2} \\
				1e3 & 82.8 & 45.2 & 41.0 \\
				\bottomrule
			\end{tabular}
		}
	\end{center}
	\caption{\textbf{Comparison of different lower-bound $C_{min}$} when clamping the denominator of linear attention into the range $[C_{min}, +\infty)$.}
	\label{tab:abl_scale}
\end{table}

To prevent dividing zeros, we clamp the denominator in Eq. (4) into the range $[C_{min}, +\infty)$.~\cref{tab:abl_scale} shows the influence of different lower-bound values $C_{min}$. When using 1e-6 of $C_{min}$, the model training fails as the activations become huge and cause loss NAN. We find 1e-1 can restrict the activation and variance in a reasonable range, leading to more stable training. When $C_{min}$ increases from 1e-1 to 1e2, the variance decrease and the training stability improves, thus the performance keeps strengthening. However, the improvement fades when using $C_{min}$ of bigger than 1e2. This is because too large $C_{min}$ will drive the whole attention map matrix close to zero and prevent distinguishing important relationships.

\section{Additional Ablation Experiments} \label{app:additional exp}
\paragraph{Linear Attention on Vanilla ViT}
\begin{table}[h]
	\begin{center}
		\setlength{\tabcolsep}{6pt}
		\footnotesize
		\resizebox{0.99\linewidth}{!}{
			\begin{tabular}{c|ccc}
				\toprule
				Model & \# Params.(M) & FLOPs(G) & Top-1 (\%) \\
				\midrule
				DeiT-T~\cite{touvron2021training} & 5.7 & 1.3 & 72.2\\ 
				DeiT-T~\cite{touvron2021training} + Enhanced Linear Attention & 5.7 & 1.3 & 72.8 \\
				\bottomrule
			\end{tabular}
		}
	\end{center}
	\caption{\textbf{Apply our proposed enhanced linear attention on the plain ViT architectures.}}
	\label{tab:abl_vanilla}
\end{table}

\cref{tab:abl_vanilla} shows the results of applying enhanced linear attention on plain architecture, i.e., DeiT. To imitate the LWA (softmax attention) + LCA (linear attention) layout of L$^2$ViT, we keep attention in half of the blocks in DeiT-T/S/B, i.e., the 1st, 3rd, 5th, 7th, 9th, 11th block, untouched. We observe that enhanced linear attention improves DeiT-Tiny by 0.6\% accuracy, while enjoying some FLOPs reduction at a small scale. The above results show that enhanced linear attention can be generalized well to plain ViT architectures.

\paragraph{Compared to Other Local Enhancements}
\begin{figure}[th]
	\begin{center}
		\includegraphics[width=0.45\textwidth]{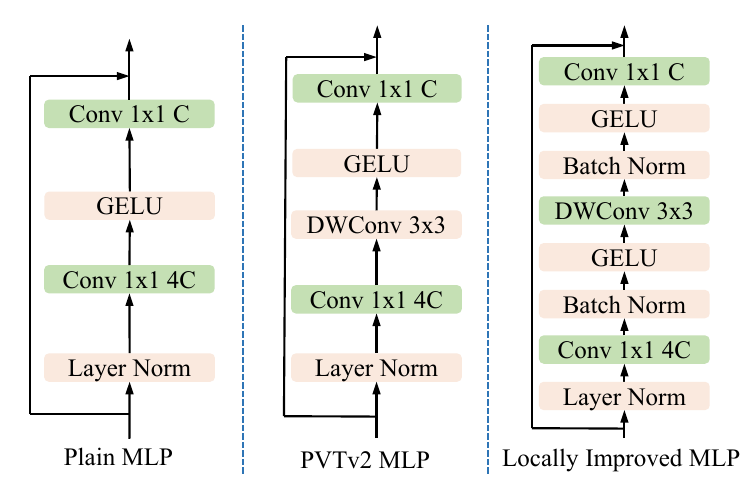}
	\end{center}
	\caption{\textbf{Comparison of different MLP layers as Plain MLP in \cite{vaswani2017attention} (left, used by L$^2$ViT), PVTv2 (middle), and locally improved MLP (right).}}
	\label{fig:abl_ole}
\end{figure}

Similar to our work, EfficientViT~\cite{cai2023efficientvit} proposes to insert a
depth-wise convolution in the MLP layer to improve the locality of feature maps generated by linear attention layers. PVTv2~\cite{wang2022pvt} also adds a 3$\times$3 depth-wise convolution to obtain more local continuity after linear spatial reduction attention. Whether LCM is advantageous in strengthening local details compared to simpler MLP with depth-wise convolution is an interesting point. It is worth noting that L$^2$ViT utilizes a plain MLP.

To confirm this, we follow PVTv2 and add a separate 3$\times$3 depth-wise convolution in plain MLP as illustrated in \cref{fig:abl_ole} middle. The results are summarized in \cref{tab:abl_ole}. However, just adding a depth-wise convolution in MLP of L$^2$ViT degrades the accuracy, which is also observed in Moat~\cite{yang2022moat}. To address this issue, we add extra normalization and activation layers between 1$\times$1 convolutions, named as locally improved MLP in~\cref{fig:abl_ole} right. Although locally improved MLP brings some improvement (0.2\% gains), it still lags behind LCM. These demonstrate that our proposed design performs more effectively in enhancing local information for linear attention output.
\begin{table}[h]
	\begin{center}
		\setlength{\tabcolsep}{16pt}
		\footnotesize
		\resizebox{0.99\linewidth}{!}{
			\begin{tabular}{l|cc}
				\toprule
				Type of Local & \#Params.(M)/  & Top-1 \\
				Enhancements  & FLOPs(G) & (\%)\\
				\toprule
				Plain MLP & 28/4.5 & 82.5  \\
				PVTv2 MLP & 29/4.7 & 81.9  \\
				Locally Improved MLP & 29/4.7 & 82.7 \\
				Plain MLP + LCM & 29/4.7 & 83.1  \\
				\bottomrule
			\end{tabular}
		}
	\end{center}
	\caption{\textbf{Top-1 accuracy of L$^2$ViT with different locally enhanced approaches.}}
	\label{tab:abl_ole}
\end{table}

\section{Local Concentration Module} \label{app:lcm}
\begin{figure}[th]
	\begin{center}
		\includegraphics[width=0.50\textwidth]{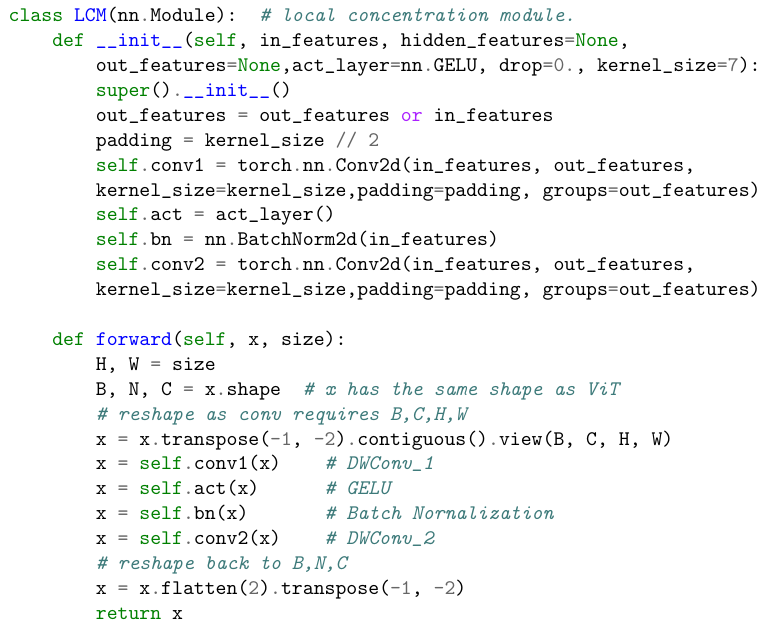}
	\end{center}
	\caption{\textbf{The Pytorch-style code for LCM.}}
	\label{fig:abl_lcm}
\end{figure}

Here we provide a detailed implementation of the local concentration module (LCM) as shown in~\cref{fig:abl_lcm}. All the source code and pre-trained models will be publicly available. Due to attention operation, we keep the feature map as $F\in \mathbb{R}^{B\times N\times C}$, $B$ is the batch size, throughout the network as Swin~\cite{liu2021swin}. However, depth-wise convolution operation requires a different feature arrangement, so we add rearrange operations to deal with this. Then two depth-wise convolutional layers are adopted to strengthen the local spatial interactions and enhance the linear attention output.

\begin{table*}[th]
	\begin{center}
		\small
		\setlength{\tabcolsep}{15pt}
		\resizebox{0.99\linewidth}{!}{
			\begin{tabular}{c|c|c|c|c}
				& \makecell{downsp. rate \\ (output size)} & L$^2$ViT-Tiny  & L$^2$ViT-Small & L$^2$ViT-Base\\
				\hline
				stem &  \makecell{4$\times$} 
				&$\begin{bmatrix}\text{conv 3$\times$3, stride 2, 48-d}\\\text{conv 3$\times$3, stride 2, 96-d}\end{bmatrix}$   
				&$\begin{bmatrix}\text{conv 3$\times$3, stride 2, 48-d}\\\text{conv 3$\times$3, stride 2, 96-d}\end{bmatrix}$    
				&$\begin{bmatrix}\text{conv 3$\times$3, stride 2, 64-d}\\\text{conv 3$\times$3, stride 2, 128-d}\end{bmatrix}$ \\
				\hline
				stage 1 & \makecell{4$\times$} &
				$\begin{bmatrix}\text{LWA, win. sz. 7$\times$7,}\\\text{LGA, LCM kernel 7$\times7$,}  \\\text{dim 96, head 3} \end{bmatrix}$ $\times$ 1   & 
				$\begin{bmatrix}\text{LWA, win. sz. 7$\times$7,}\\\text{LGA, LCM kernel 7$\times7$,} \\\text{dim 96, head 3} \end{bmatrix}$ $\times$ 1    & 
				$\begin{bmatrix}\text{LWA, win. sz. 7$\times$7,}\\\text{LGA, LCM kernel 7$\times7$,} \\\text{dim 128, head 4} \end{bmatrix}$ $\times$ 1   \\
				\hline
				\multirow{4}{*}{stage 2}  & \multirow{5}{*}{\makecell{8$\times$}} & conv 2$\times$2, stride 2, 192-d & conv 2$\times$2, stride 2, 192-d & conv 2$\times$2, stride 2, 256-d  \\
				\cline{3-5}
				& & $\begin{bmatrix}\text{LWA, win. sz. 7$\times$7,}\\\text{LGA, LCM kernel 7$\times7$,} \\\text{dim 192, head 6} \end{bmatrix}$ $\times$ 1   & 
				$\begin{bmatrix}\text{LWA, win. sz. 7$\times$7,}\\\text{LGA, LCM kernel 7$\times7$,} \\\text{dim 192, head 6} \end{bmatrix}$ $\times$ 1    & 
				$\begin{bmatrix}\text{LWA, win. sz. 7$\times$7,}\\\text{LGA, LCM kernel 7$\times7$,} \\\text{dim 256, head 8} \end{bmatrix}$ $\times$ 1   \\
				\hline
				\multirow{4}{*}{stage 3}  & \multirow{5}{*}{\makecell{16$\times$}}  & conv 2$\times$2, stride 2, 384-d & conv 2$\times$2, stride 2, 384-d & conv 2$\times$2, stride 2, 512-d  \\
				\cline{3-5}
				& & $\begin{bmatrix}\text{LWA, win. sz. 7$\times$7,}\\\text{LGA, LCM kernel 7$\times7$,} \\\text{dim 384, head 12} \end{bmatrix}$ $\times$ 3   & 
				$\begin{bmatrix}\text{LWA, win. sz. 7$\times$7,}\\\text{LGA, LCM kernel 7$\times7$,} \\\text{dim 384, head 12} \end{bmatrix}$ $\times$ 9    & 
				$\begin{bmatrix}\text{LWA, win. sz. 7$\times$7,}\\\text{LGA, LCM kernel 7$\times7$,} \\\text{dim 512, head 16} \end{bmatrix}$ $\times$ 9   \\
				\hline
				\multirow{4}{*}{stage 4} & \multirow{5}{*}{\makecell{32$\times$}}  & conv 2$\times$2, stride 2, 768-d & conv 2$\times$2, stride 2, 768-d & conv 2$\times$2, stride 2, 1024-d \\
				\cline{3-5}
				& & $\begin{bmatrix}\text{LWA, win. sz. 7$\times$7,}\\\text{LGA, LCM kernel 7$\times7$,} \\\text{dim 768, head 24} \end{bmatrix}$ $\times$ 1   & $\begin{bmatrix}\text{LWA, win. sz. 7$\times$7,}\\\text{LGA, LCM kernel 7$\times7$,} \\\text{dim 768, head 24} \end{bmatrix}$ $\times$ 1   & 
				$\begin{bmatrix}\text{LWA, win. sz. 7$\times$7,}\\\text{LGA, LCM kernel 7$\times7$,} \\\text{dim 1024, head 32} \end{bmatrix}$ $\times$ 1   \\
		\end{tabular}}
	\end{center}
	\normalsize
	\caption{\textbf{Detailed architecture configurations.} }
	\label{tab:arc detail}
\end{table*}

\section{Limitations}
While our proposed local concentration module enhances the linear attention to a large extent, we notice that the dispersive attention in deeper layers like layer 12 in Figure 2. may not be compensated by convolution since they show global patterns instead of local patterns. We think that developing a specific concentration module for deep layers or considering applying vanilla attention directly will be an interesting future direction to improve our work. 

Some works~\cite{ali2021xcit, ding2022davit} explore channel attention building channel-to-channel interactions instead of patch-to-patch interactions and maintaining linear complexity. Both linear attention and channel attention calculate the multiplication of key and value first. They differ in several aspects: First, linear attention still models the spatial relationship, while the latter focuses on channel dependency. Second, the removal of softmax decouples the computational order of attention in Eq.~4. Thus linear attention can dynamically choose whether to multiply $Q$ and $K$ first with the complexity of $O(N^2C)$ if $C$ is large or multiply $K$ and $V$ first with $O(NC^2)$ when $N$ is large to maintain optimal efficiency according to the input size. Third, softmax increases computational overheads and is inefficient in many practical applications. Last but not least, we empirically show linear attention achieves superior performance in Tab.~8.

\section{Model Configurations} \label{app:model config}
Table~\ref{tab:arc detail} shows the detailed model configurations for L$^2$ViT-Tiny/Small/Base. Unlike the non-overlapping patchify stem in Swin~\cite{liu2021swin}, we adopt a two-layer convolutional stem to extract more important local structure information for each patch. In $i$th stage, we alternatively arrange $N_i$ LWA and $N_i$ LGA, total 2$N_i$ blocks. In this way, LWA first models short-range interactions, then LGA constructs global patch-to-patch relationships. LGA can reinforce the holistic perception of features encoded by LWA to boost the expressivity of the model. Both LGA and LWA apply MLP (expansion ratio of 4), the same as the DeiT~\cite{touvron2021training}, to model channel relationships. Besides, Both LGA and LWA in all stages adopt CPE (kernel size 3$\times$3) as position embedding as CPE is more friendly to various input resolutions.

\end{document}